\documentclass[runningheads]{llncs}
\usepackage{comment}
\usepackage[T1]{fontenc}
\usepackage{hyperref}
\usepackage{xcolor}
\usepackage{booktabs}
\usepackage{graphicx}
\usepackage{array}
\newcolumntype{P}[1]{>{\centering\arraybackslash}p{#1}}
\usepackage{makecell}
\usepackage[backend=biber,style=numeric,sorting=none,maxcitenames=2]{biblatex}
\usepackage{amsmath}   
\usepackage{amssymb}   
\usepackage{multicol}
\usepackage{multirow}

\usepackage{color}
\makeatletter

\renewcommand\subsection{%
  \@startsection{subsection}{2}{\z@}%
    {6pt}
    {3pt}
    {\normalfont\normalsize\bfseries}%
}

\renewcommand\subsubsection{%
  \@startsection{subsubsection}{3}{\z@}%
    {4pt}
    {2pt}
    {\normalfont\normalsize\itshape}%
}

\makeatother

\usepackage{enumitem}
\setlist[itemize]{topsep=2pt, partopsep=0pt, itemsep=1pt, parsep=0pt}
\setlist[enumerate]{topsep=2pt, partopsep=0pt, itemsep=1pt, parsep=0pt}

\begin{document}
\title{X-MAP: eXplainable Misclassification Analysis and Profiling for Spam and Phishing Detection}
\titlerunning{X-MAP: eXplainable Misclassification Analysis and Profiling}
%
\author{Qi Zhang\inst{1}\orcidID{0000-0002-3607-3258} \and
Dian Chen\inst{1}\orcidID{0009-0000-7641-454X} \and
Lance M.~Kaplan\inst{2}\orcidID{0000-0002-3627-4471} \and
Audun J{\o}sang\inst{3}\orcidID{0000-0001-6337-2264} \and
Dong Hyun Jeong\inst{4}\orcidID{0000-0001-5271-293X} \and
Feng Chen\inst{5}\orcidID{0000-0002-4508-5963} \and
Jin-Hee Cho\inst{1}\orcidID{0000-0002-5908-4662}}

\authorrunning{Zhang et al.}
\institute{Virginia Tech, USA \and
U.S. Army DEVCOM Army Research Laboratory, USA \and
University of Oslo, Norway \and
University of the District of Columbia, USA \and 
University of Texas at Dallas, USA}

\maketitle              
\begin{abstract}
Misclassifications in spam and phishing detection are very harmful, as false negatives expose users to attacks while false positives degrade trust. Existing uncertainty-based detectors can flag potential errors, but possibly be deceived and offer limited interpretability. This paper presents X-MAP, an e\underline{X}plainable \underline{M}isclassification \underline{A}nalysis and \underline{P}rofilling framework that reveals topic-level semantic patterns behind model failures. X-MAP combines SHAP-based feature attributions with non-negative matrix factorization to build interpretable topic profiles for reliably classified spam/phishing and legitimate messages, and measures each message’s deviation from these profiles using Jensen--Shannon divergence. Experiments on SMS and phishing datasets show that misclassified messages exhibit at least two times larger divergence than correctly classified ones. As a detector, X-MAP achieves up to 0.98 AUROC and lowers the false-rejection rate at 95\% TRR to 0.089 on positive predictions. When used as a repair layer on base detectors, it recovers up to 97\% of falsely rejected correct predictions with moderate leakage. These results demonstrate X-MAP’s effectiveness and interpretability for improving spam and phishing detection.
\end{abstract}

\section{Introduction}

Digital communication platforms, such as email and Short Message Service (SMS), increasingly expose users to spam and phishing attacks, which remain among the most prevalent and costly cyber threats. Phishing was the most frequently reported internet crime in 2023~\cite{FBI2024IC3Report}, and SMS-based scams resulted in over \$470 million in losses in 2024~\cite{FTC2025TextScams2024}. Although machine learning (ML) models have shown strong performance in spam and phishing detection~\cite{gutierrez2020email, nair2025phishemailllm}, they still produce unavoidable misclassifications. Such errors have direct consequences: false negatives expose users to harmful content, while false positives undermine user experience and trust. Understanding why ML models fail is therefore as important as improving accuracy. Misclassifications often come from subtle feature patterns, dataset biases, or mismatched decision boundaries that are not captured by traditional performance metrics. Interpreting these failures can reveal structural weaknesses in the detection pipeline and provide actionable insights for improving robustness and transparency.

This work analyzes misclassification behaviors instead of optimizing classification accuracy, aiming to identify interpretable indicators and text features linked to incorrect predictions and to provide insights that advance transparent, trustworthy, and security-enhancing machine learning applications.

The \textbf{key contributions} of this work are:
\begin{itemize}
    \item We propose e\underline{X}plainable \underline{M}isclassification \underline{A}nalysis and \underline{P}rofiling (X-MAP), a framework that integrates \underline{SH}apley \underline{A}dditive ex\underline{P}lanations (SHAP) with topic modeling to interpret model behavior at topic-level resolutions, offering fine-grained explanations for why specific messages are misclassified.

    \item We introduce a mechanism to cluster top-ranked SHAP features using \textit{non-negative matrix factorization} (NMF) to form higher-level interpretable semantic topics. We separately analyze positive and negative SHAP values as they behave differently under positive and negative predictions.
    
    \item We construct \textit{group-level topic prototypes} for correct classification groups and compare individual messages to these profiles using distance-based measures (i.e., Jensen, Shannon divergence), providing a quantitative view of how each message aligns with known correct classification patterns.

\end{itemize}

\section{Related Work}

\textbf{Uncertainty-Based Misclassification Detection.}  
A substantial body of work detects misclassifications using uncertainty signals derived from model predictions. Hendrycks and Gimpel~\cite{hendrycks2017baseline} demonstrated that prediction confidence provides a strong baseline for identifying both misclassified and out-of-distribution samples. Qiu and Miikkulainen~\cite{qiu2022detecting} proposed RED, which leverages Gaussian Processes to model uncertainty in a classifier-agnostic manner. Granese et al.~\cite{granese2021doctor} introduced DOCTOR, a softmax-based accept, reject mechanism that rejects decisions deemed uncertain. For Transformer-based NLP models, Vazhentsev et al.~\cite{vazhentsev2022uncertainty} showed that Monte Carlo dropout and deep ensembles improve the detection of erroneous predictions. Dadalto et al.~\cite{gomes2023rel-u} presented REL-U, which captures inter-sample relationships beyond entropy and often outperforms traditional uncertainty measures. In intrusion detection, Alyahya et al.~\cite{alyahya2024toward} combined RNN and SVM architectures to reduce misclassifications. While these methods effectively flag potential errors, they generally do not explain \emph{why} models fail, offering limited interpretability into the underlying causes.

\noindent \textbf{Data-Centric Misclassification Identification.}  
Another direction focuses on identifying potentially misclassified samples through data-level signals. Zhu et al.~\cite{zhu2023openmix} proposed OpenMix, a data-augmentation technique that mixes inputs with synthetic outliers to improve misclassification and OOD detection, though its performance is sensitive to the quality of generated outliers and is less evaluated on text data. Naser~\cite{naser2025failure} surveyed methods for learning under noisy or mislabeled data, emphasizing robustness through error-pattern analysis. Klie et al.~\cite{klie2023annotation} detected annotation errors by measuring consistency and uncertainty across models, revealing mislabeled instances. Du et al.~\cite{du2025balancing} introduced a hierarchical severity metric to quantify the impact of misclassifications, enabling more informed error prioritization. Spackman et al.~\cite{spackman2023identifying} outlined strategies for identifying and mitigating misclassifications in large-scale economic classification systems. Although these data-centric approaches help improve data quality and model robustness, they largely focus on detecting or filtering problematic samples and provide limited insight into the semantic or structural reasons behind misclassifications.



\section{Proposed Approach: X-MAP}

\begin{figure}[t]
\centering
\includegraphics[width=0.8\textwidth]{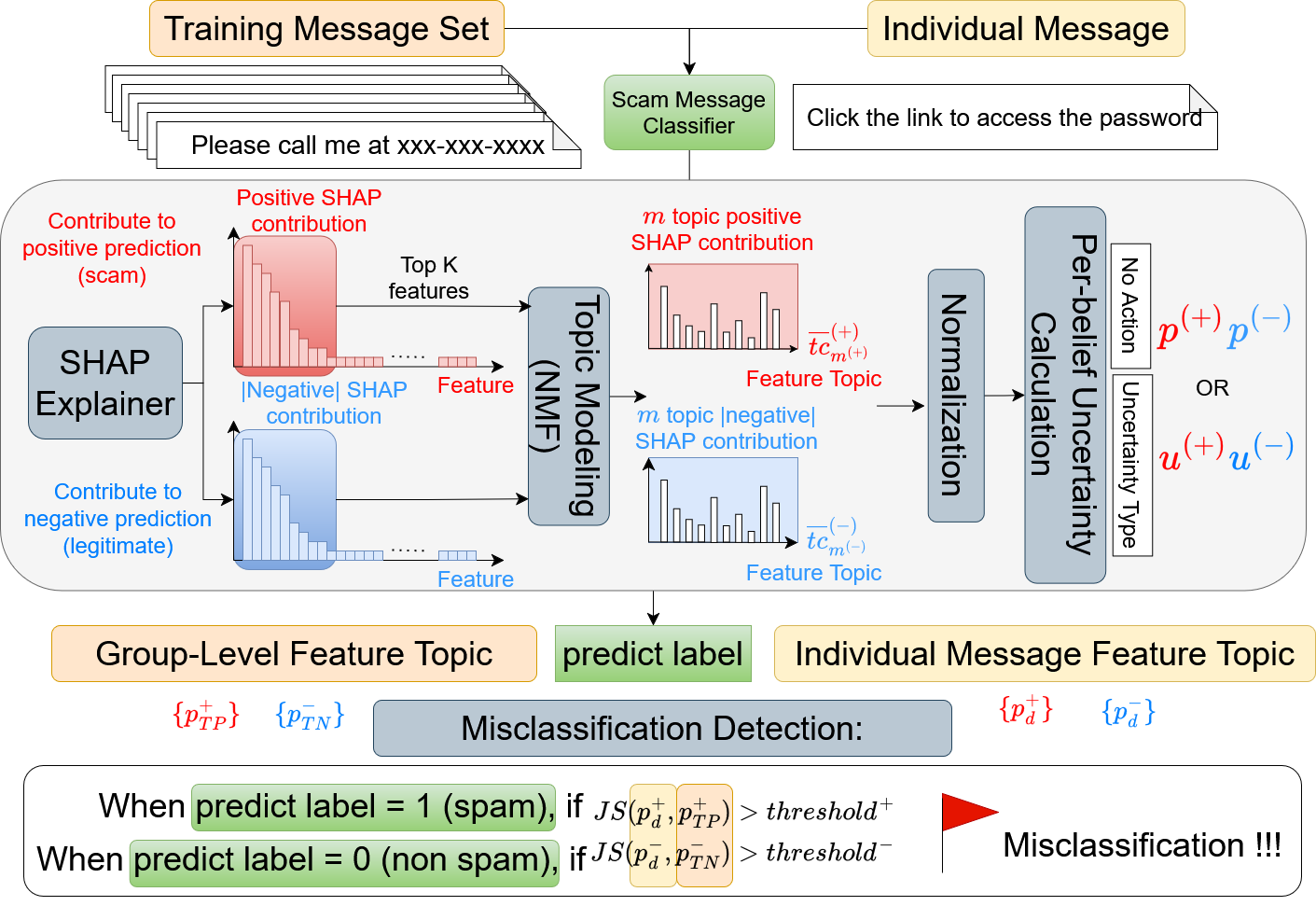}
\vspace{-3mm}
\caption{X-MAP for explainable misclassification analysis in spam/phishing detection.
}
\label{fig:framework-overview}
\end{figure}

We propose X-MAP, a framework that explains and detects misclassifications through topic-level analysis of SHAP explanations. As shown in Fig.~\ref{fig:framework-overview}, X-MAP comprises four stages: (1) train a binary classifier for spam/phishing detection; (2) compute SHAP~\cite{SHAP} values for each feature in each message pair to capture their contributions to positive and negative classes; (3) apply nonnegative matrix factorization to the SHAP matrices to derive interpretable topics and group-level profiles for true positives (TP) and true negatives (TN); and (4) aggregate each message’s SHAP values into topic distributions and measure their divergence from the corresponding reliable group using Jensen--Shannon (JS) divergence, where larger divergence signals higher misclassification likelihood.

\subsection{Stage 1: Classifier and Feature Representation}
\label{sec:token feature representation}

\noindent We consider three standard text classifiers: Logistic Regression (LogReg), Support Vector Machine (SVM), and Na\"{\i}ve Bayes (NB)~\cite{murphy2012mlpp}. All models are trained on the same multiview feature space consist of 7000 TF-IDF word $n$-grams, 3000 TF-IDF phrase $n$-grams, and 17 handcrafted structural features (e.g., character and word counts, average word length, number of digits). Each message is represented as a $(7000 + 3000 + 17)$-dimensional vector.

\subsection{Stage 2: SHAP-Based Feature Ranking}
\label{sec:shap-ranking}

\noindent Let $\boldsymbol{\phi}^{(i)} \in \mathbb{R}^d$ be the SHAP vector for message $i$. We split it into two nonnegative “supports” for the positive and negative classes:
\begin{equation}
S^{(+)}_{ij} = \max(\phi^{(i)}_j, 0), \qquad
S^{(-)}_{ij} = \max(-\phi^{(i)}_j, 0).
\end{equation}
Here $S^{(\pm)}_{ij}$ reflects how feature $j$ supports positive (spam/phishing) or negative (legitimate) decisions for message $i$. We use \texttt{LinearExplainer} for LogReg and \texttt{KernelExplainer} for SVM and NB to match each model's decision function.

\noindent Within each class ($+$ or $-$) and feature $j$, we measure global importance by:
\begin{equation}
\mathrm{presence}^{(\pm)}_j = \frac{1}{n} \sum_{i=1}^{n} \mathbb{I}\!\left[S^{(\pm)}_{ij} \neq 0\right], \; \; 
\mathrm{cond\_mean}^{(\pm)}_j = \frac{\sum_{i=1}^{n} S^{(\pm)}_{ij}}{\sum_{i=1}^{n} \mathbb{I}\!\left[S^{(\pm)}_{ij} \neq 0\right]},
\end{equation}
where $\mathrm{presence}^{(\pm)}_j$ measures how often feature $j$ contributes to class $(\pm)$ decisions, and $\mathrm{cond\_mean}^{(\pm)}_j$ captures its average contribution when feature $j$ is active. We define the ranking score by:
\begin{equation}
r^{(\pm)}_j = \mathrm{cond\_mean}^{(\pm)}_j \cdot \sqrt{\max\!\big(\mathrm{presence}^{(\pm)}_j,\, \tau_p\big)},
\end{equation}
where $\tau_p$ is a small presence floor to prevent rare features from dominating the ranking. To balance feature dominations, we allocate quota ratios $\{\rho_0, \rho_1, \rho_2\}$ to select $K$ top-ranked features from three feature types for each class.

Let $\mathcal{F}^{(\pm)}_{\text{top}}$ denote the set of top-ranked features selected for the positive ($+$) or negative ($-$) polarity, representing feature patterns most indicative of spam/phishing or legitimate messages. Using these features, we construct the nonnegative SHAP matrices by:
\begin{equation}
\mathbf{X}^{(\pm)} \in \mathbb{R}^{\,n \times K^{(\pm)}}_{\ge 0}, \qquad 
X^{(\pm)}_{if} = S^{(\pm)}_{if} \;\; \text{for } f \in \mathcal{F}^{(\pm)}_{\text{top}},
\end{equation}
where $\mathbf{X}^{(\pm)}$ collects SHAP contributions of all selected features for classes $(\pm)$ across all $n$ messages, and $K^{(\pm)} = |\mathcal{F}^{(\pm)}_{\text{top}}|$ is the number of selected features for that class. These matrices serve as the input to topic modeling. 

\subsection{Stage 3: Topic Modeling and Feature Patterns} 

\noindent \textbf{Topic Assignment with NMF.}  We apply \textit{Nonnegative Matrix Factorization} (NMF)~\cite{lee2000nmf} to each SHAP matrix, decomposing $\mathbf{X}^{(\pm)}$ as: 
\begin{equation}
\mathbf{X}^{(\pm)} \approx \mathbf{W}^{(\pm)} \mathbf{H}^{(\pm)},
\end{equation}
where $\mathbf{W}^{(\pm)} \in \mathbb{R}_{\ge 0}^{n \times M^{(\pm)}}$ is the message--topic matrix, $\mathbf{H}^{(\pm)} \in \mathbb{R}_{\ge 0}^{M^{(\pm)} \times K^{(\pm)}}$ is the topic--feature matrix, and $M^{(\pm)}$ is the number of latent topics for class $(\pm)$. Each feature $j$ (indexed within the selected feature set) is assigned to the topic with the largest weights in $\mathbf{H}^{(\pm)}$:
\begin{equation}
\mathrm{topic}^{(\pm)}(j)
= \arg\max_{m \in \{1,\dots,M^{(\pm)}\}} H^{(\pm)}_{m j}.
\end{equation}
For message $i$, we then sum up SHAP contributions across all features belonging to topic $m$ to obtain the topic-level contribution by:
\begin{equation}
tc^{(\pm)}_{i m}
= \sum_{j:\,\mathrm{topic}^{(\pm)}(j)=m} X^{(\pm)}_{i j},
\end{equation}
which represents how strongly message $i$ expresses topic $m$ under class $(\pm)$.

\noindent \textbf{Group-Level Feature Topic Distributions.}  We partition correctly classified messages into true positives (TP) and true negatives (TN) groups, denoted by $D_{\mathrm{TP}}$ and $D_{\mathrm{TN}}$. For each group $g \in \{\mathrm{TP},\mathrm{TN}\}$ and topic $m$ in class $(\pm)$, we compute the average topic contribution
\begin{equation}
\overline{tc}^{(\pm)}_{g,m}
= \frac{1}{|D_g|} \sum_{i \in D_g} tc^{(\pm)}_{i,m},
\end{equation}
where $\overline{tc}^{(\pm)}_{g,m}$ reflects how strongly group $g$ expresses topic $m$. We then normalize these averages to obtain the group-level \emph{feature topic distribution}
\begin{equation}
p^{(\pm)}_{g,m}
= \frac{\overline{tc}^{(\pm)}_{g,m}}{\sum_{m} \overline{tc}^{(\pm)}_{g,m}},
\end{equation}
which serves as the reliable topic profile for positive (spam/phishing) and negative (legitimate) predictions.

\noindent \textbf{Group-Level Topic-Specific Uncertainty Measures.}
\noindent We compute group-level \textit{topic-specific uncertainty} measures from the distribution $p^{(\pm)}_{g,m}$. \textit{Vacuity}~\cite{jsang2018subjective} reflects the strength of evidence for a topic, with higher values indicating weaker support. \textit{Dissonance}~\cite{jsang2018subjective} measures the conflict among similarly probable topics, revealing inconsistency in the explanation space. \textit{Aleatory} uncertainty captures entropy among related topics, indicating how dispersed or ambiguous the contributions are for group $g$. We further apply topic-wise variants of DOCTOR~\cite{granese2021doctor}, REL-U~\cite{gomes2023rel-u}, and ODIN~\cite{liang2018enhancing} by evaluating their uncertainty definitions on each topic probability; formulas follow the original works and are omitted for brevity.

\subsection{Stage 4: Topic-Based Misclassification Scoring}
\label{sec:misclassification-scoring}

\noindent Given a classifier and the topic representations, X-MAP assigns the message $d$ a misclassification score based on the Jensen--Shannon (JS) divergence~\cite{lin2002divergence}. Let $p^{(\pm)}_d$ denote the topic distribution of message $d$ under class $(\pm)$, and let $p^{(\pm)}_{g}$ be the corresponding group-level reliable topic profile for group $g \in \{\mathrm{TP}, \mathrm{TN}\}$. The misclassification score is given by:
\begin{equation}
score_d = \mathrm{JS}\!\left( p^{(\pm)}_d,\; p^{(\pm)}_{g} \right), \qquad
g =
\begin{cases}
\mathrm{TP}, & \text{if } \hat{y}_d = 1,\\[4pt]
\mathrm{TN}, & \text{if } \hat{y}_d = 0,
\end{cases}
\end{equation}
where $\hat{y}_d$ is the classifier’s predicted label. Higher JS divergence indicates greater deviation from the reliable group and thus higher misclassification likelihood.

Correctly classified messages align with their group-level topic profiles, whereas misclassified ones deviate, as shown in Section~\ref{sec:distance assumption} by their larger $score_d$ values. The same approach applies to topic-specific uncertainty (vacuity, dissonance, etc.) by normalizing these values and computing their JS divergence from group-level uncertainty profiles. The resulting $score_d$ serves as a misclassification indicator or as a ``repair layer'' alongside existing UQ detector.

\subsection{Repair Layer}
\label{sec: repair layer design}

\begin{figure}[t]
\centering
\includegraphics[width=0.9\textwidth]{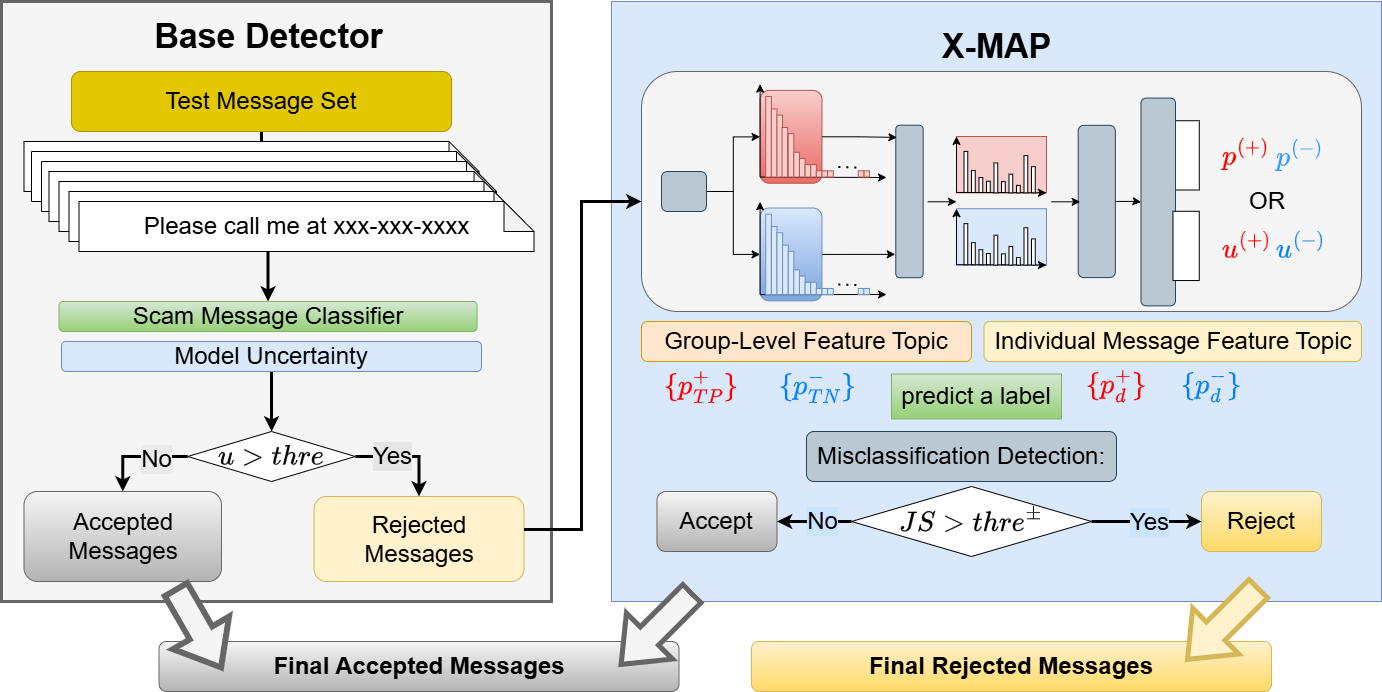}
\vspace{-3mm}
\caption{X-MAP Re-evaluation for Error Recovery: This enables recovery of correctly classified samples while keeping error leakage low.
}

\label{fig:repair-layer}
\end{figure}

\noindent Traditional UQ-based detectors flag potential misclassifications using predictive uncertainty, whereas X-MAP analyzes feature-level behavior and thus complements UQ signals. We exploit this complementarity by using X-MAP as a repair layer on top of a UQ detector, as illustrated in Fig.~\ref{fig:repair-layer}.

Given a base classifier, we first apply several state-of-the-art UQ indicators, including Vacuity and Dissonance~\cite{Josang18}, Entropy~\cite{renyi1961entropy}, DOCTOR~\cite{granese2021doctor}, ODIN~\cite{liang2017odin}, and REL-U~\cite{gomes2023rel-u}, to obtain a rejection set at a fixed true rejection rate $TRR_{fix}$. X-MAP is then applied to the rejected messages to identify those whose topic patterns align with the corresponding reliable group (TP or TN), allowing them to be re-accepted. This two-stage process increases coverage by recovering correct predictions while keeping error leakage low, demonstrating utility beyond UQ-based detectors at the same $TRR_{fix}$.

\section{Experiment Setup}

We use two text-based \textbf{Datasets}: the \emph{UCI SMS Spam Collection}~\cite{sms_spam_collection_228} and a \emph{Phishing Emails} dataset combining Ling~\cite{LingEmail} and Nigerian Fraud~\cite{NigerianFraud} emails. We augment both with an LLM to balance labels and remove stopwords and rare tokens. Instead of the standard 80/20 split, we adopt a 50/50 train, test split to increase evaluation samples, as misclassifications are infrequent and the classifiers perform well even with less training data.

\noindent \textbf{Parameters} include quota ratios $\{\rho_0,\rho_1,\rho_2\}=\{0.65,0.3,0.05\}$ for feature allocation. We select the top $K=200$ features with a presence floor $\tau_p=0.05$. X-MAP uses $m=10$ latent topics and considers the $k=2$ most similar topics for each when calculating aleatory uncertainty. The fixed true reject rate is set to $TRR_{fix}=95\%$ when X-MAP is applied to rejected messages.

\noindent \textbf{Metrics} used to evaluate X-MAP include: \textbf{(1) AUROC}, summarizing the trade-off between the true rejection rate (TRR) and false rejection rate (FRR); \textbf{(2) FRR at 95\% TRR}, the share of correctly classified samples rejected when TRR is at 95\%; \textbf{(3) Recovery Rate} ($RecovR$), the fraction of a base detector’s false rejections that X-MAP restores; and \textbf{(4) Error Leakage Rate} ($LeakR$), the fraction of misclassified samples correctly rejected by the base detector but mistakenly re-accepted by X-MAP.

\section{Experiment Results \& Analyses} 

\subsection{Distance Analysis of Correct Classified vs. Misclassified Samples} 
\label{sec:distance assumption}
\begin{table}[t]
\centering
\caption{Jensen--Shannon divergence between individual messages (IM) and group topic profiles (GP) on the SMS dataset}
\label{tab:jsd_distance_topic_distribution}
\scriptsize
\begin{tabular}{|l|l|c|c|c|c|}
\hline
\textbf{CFR} 
& \textbf{GP type} 
& \textbf{TP$\to$TP GP} 
& \textbf{FP$\to$TP GP} 
& \textbf{TN$\to$TN GP} 
& \textbf{FN$\to$TN GP} \\
\hline
\multirow{8}{*}{LogReg}
& \textbf{Ori.} & \textbf{0.2051 $\pm$ 0.1405} & 0.5490 $\pm$ 0.1440 & \textbf{0.0543 $\pm$ 0.0524} & 0.1213 $\pm$ 0.0518\\
\cline{2-6}
& \textbf{Vac.} & \textbf{0.1946 $\pm$ 0.1281} & 0.5010 $\pm$ 0.1180 & \textbf{0.0536 $\pm$ 0.0500} & 0.1207 $\pm$ 0.0521\\
\cline{2-6}
& \textbf{Diss.} & \textbf{0.2213 $\pm$ 0.1320} & 0.5336 $\pm$ 0.1164 & \textbf{0.0603 $\pm$ 0.0634} & 0.1467 $\pm$ 0.0658\\
\cline{2-6}
& \textbf{Ale.} & \textbf{0.0093 $\pm$ 0.0093} & 0.0600 $\pm$ 0.0338 & \textbf{0.0090 $\pm$ 0.0136} & 0.0271 $\pm$ 0.0149\\
\cline{2-6}
& \textbf{D${\alpha}$} & \textbf{0.0050 $\pm$ 0.0050} & 0.0202 $\pm$ 0.0125 & \textbf{0.0006 $\pm$ 0.0014} & 0.0015 $\pm$ 0.0011 \\
\cline{2-6}
& \textbf{D${\beta}$} & \textbf{0.0015 $\pm$ 0.0017} & 0.0105 $\pm$ 0.0115 & \textbf{0.0001 $\pm$ 0.0003}& 0.0003 $\pm$ 0.0003\\
\cline{2-6}
& \textbf{ODIN} & \textbf{0.0026 $\pm$ 0.0024} & 0.0128 $\pm$ 0.0118 & \textbf{0.0014 $\pm$ 0.0007}& 0.0014 $\pm$ 0.0008\\
\cline{2-6}
& \textbf{REL-U} & \textbf{0.1753 $\pm$ 0.1993} & 0.4454 $\pm$ 0.1400 & \textbf{0.1675 $\pm$ 0.0798} & 0.1744 $\pm$ 0.0750 \\
\hline
\multirow{8}{*}{SVM}
& \textbf{Ori.} & \textbf{0.1772 $\pm$ 0.1162} & 0.4571 $\pm$ 0.1086 & \textbf{0.1063 $\pm$ 0.0816} & 0.1923 $\pm$ 0.1052\\
\cline{2-6}
& \textbf{Vac.} & \textbf{0.1674 $\pm$ 0.1133} & 0.3643 $\pm$ 0.1348 & \textbf{0.1403 $\pm$ 0.0878} & 0.2470 $\pm$ 0.1095\\
\cline{2-6}
& \textbf{Diss.} & \textbf{0.1615 $\pm$ 0.1247} & 0.3928 $\pm$ 0.1068 & \textbf{0.1016 $\pm$ 0.1001} & 0.2274 $\pm$ 0.1257\\
\cline{2-6}
& \textbf{Ale.} & \textbf{0.0530 $\pm$ 0.0294} & 0.1337 $\pm$ 0.0530 & \textbf{0.0465 $\pm$ 0.0428} & 0.0615 $\pm$ 0.0540\\
\cline{2-6}
& \textbf{D${\alpha}$} & \textbf{0.0046 $\pm$ 0.0046} & 0.0240 $\pm$ 0.0143 & \textbf{0.0018 $\pm$ 0.0035} & 0.0035 $\pm$ 0.0035\\
\cline{2-6}
& \textbf{D${\beta}$} & \textbf{0.0017 $\pm$ 0.0026} & 0.0209 $\pm$ 0.0188 & \textbf{0.0010 $\pm$ 0.0019} & 0.0016 $\pm$ 0.0021\\
\cline{2-6}
& \textbf{ODIN} & \textbf{0.0028 $\pm$ 0.0025} & 0.0191 $\pm$ 0.0166 & \textbf{0.0028 $\pm$ 0.0028} & 0.0039 $\pm$ 0.0032\\
\cline{2-6}
& \textbf{REL-U} & \textbf{0.1030 $\pm$ 0.1291} & 0.3786 $\pm$ 0.1609 & 0.4080 $\pm$ 0.0849 & \textbf{0.3361 $\pm$ 0.1871}\\
\hline
\multirow{8}{*}{\makecell{Na\"{\i}ve \\ Bayes}}
& \textbf{Ori.} & \textbf{0.2413 $\pm$ 0.1030} & 0.6062 $\pm$ 0.2357 & \textbf{0.1204 $\pm$ 0.0605} & 0.1997 $\pm$ 0.1599\\
\cline{2-6}
& \textbf{Vac.} & \textbf{0.2332 $\pm$ 0.0891} & 0.2912 $\pm$ 0.2022 & \textbf{0.2920 $\pm$ 0.0765} & 0.3001 $\pm$ 0.0876\\
\cline{2-6}
& \textbf{Diss.} & \textbf{0.2404 $\pm$ 0.0978} & 0.3397 $\pm$ 0.2145 & \textbf{0.2159 $\pm$ 0.0444} & 0.2524 $\pm$ 0.0863\\
\cline{2-6}
& \textbf{Ale.} & \textbf{0.0257 $\pm$ 0.0182} & 0.0746 $\pm$ 0.0375 & \textbf{0.0661 $\pm$ 0.0396} & 0.0862 $\pm$ 0.0375\\
\cline{2-6}
& \textbf{D${\alpha}$} & \textbf{0.0041 $\pm$ 0.0050} & 0.0341 $\pm$ 0.0208 & \textbf{0.0029 $\pm$ 0.0043} & 0.0068 $\pm$ 0.0114 \\
\cline{2-6}
& \textbf{D${\beta}$} & \textbf{0.0011 $\pm$ 0.0042} & 0.0301 $\pm$ 0.0228 & \textbf{0.0037 $\pm$ 0.0058} & 0.0058 $\pm$ 0.0098 \\
\cline{2-6}
& \textbf{ODIN} & \textbf{0.0020 $\pm$ 0.0044} & 0.0309 $\pm$ 0.0236 & \textbf{0.0029 $\pm$ 0.0045} & 0.0078 $\pm$ 0.0109 \\
\cline{2-6}
& \textbf{REL-U} & \textbf{0.1530 $\pm$ 0.1106} & 0.3110 $\pm$ 0.1893 & NA & NA\\
\hline
\end{tabular}

\vspace{2pt}
\raggedright
CFR = Classifier; TP$\to$TP = distance from TP IMs to TP GP; FP$\to$TP = FP IMs to TP GP; TN$\to$TN = TN IMs to TN GP; FN$\to$TN = FN IMs to TN GP. LogReg = Logistic Regression, Na\"{\i}ve B. = Na\"{\i}ve Bayes. NA means no group-level feature topic distribution is available due to the base classifier's performance, so there is no way to separate misclassification
\end{table}

\noindent X-MAP builds on the intuition that correctly classified messages align with their reliable group’s topic profile, whereas misclassified ones do not. Table~\ref{tab:jsd_distance_topic_distribution} confirms this pattern holds consistently: true positives show lower JS divergence to the TP profile than false positive smaples, and true negatives show the same trend relative to TN. The only exception is REL-U on negative predictions, where the separation is weaker, suggesting that REL-U is less suitable as a topic-level measure for our setting. Results are shown for the SMS dataset, with the phishing dataset demonstrating the same qualitative pattern.

\begin{table}[t]
\centering
\caption{Performance of X-MAP on SMS Dataset}
\label{tab:auroc_and_frr_topic_specific_uncertainty}
\scriptsize
\setlength{\tabcolsep}{2pt}  
\begin{tabular}{|c|l|c|c|c|c|c|c|c|c|c|}
\hline
\multirow{2}{*}{\textbf{CFR}} 
& \multirow{2}{*}{\textbf{Metric}} 
& \multicolumn{8}{c|}{\textbf{Topic Representation}} 
& \multirow{2}{*}{\makecell{\textbf{UQ-D}\\}} \\
\cline{3-10}
& & Ori. & Ale. & Vac. & Diss. & REL-U & $D_\alpha$ & $D_\beta$ & ODIN & \\
\hline

\multirow{4}{*}{LogReg} 
& AUROC (P)        & 0.9409 & \textbf{0.9763} & 0.9470 & 0.9474 & 0.8283 & 0.8954 & 0.8545 & 0.8545 & 0.9367 \\
& FRR@95\%TRR (P)  & 0.3988 & \textbf{0.0895} & 0.3923 & 0.2301 & 0.2321 & 0.5900 & 0.6054 & 0.6054 & 0.6304 \\
& AUROC (N)        & 0.8807 & 0.8253 & 0.8802 & 0.8711 & 0.5252 & 0.8536 & 0.8041 & 0.8041 & \textbf{0.9369} \\
& FRR@95\%TRR (N)  & 0.3588 & 0.6186 & 0.3574 & 0.3971 & 0.9397 & 0.4130 & 0.5515 & 0.5515 & \textbf{0.2247} \\
\hline

\multirow{4}{*}{SVM} 
& AUROC (P)        & 0.9574 & 0.9264 & 0.8570 & 0.9063 & 0.9057 & 0.9316 & 0.9090 & 0.9090 & \textbf{0.9734} \\
& FRR@95\%TRR (P)  & 0.2350 & 0.3284 & 0.3902 & 0.2920 & 0.3952 & 0.3547 & 0.4554 & 0.4554 & \textbf{0.1176} \\
& AUROC (N)        & 0.7890 & 0.5722 & 0.7741 & 0.8027 & 0.4008 & 0.7538 & 0.6323 & 0.6323 & \textbf{0.9028} \\
& FRR@95\%TRR (N)  & 0.8512 & 0.9545 & 0.8531 & 0.7351 & 0.9953 & 0.7122 & 0.8589 & 0.8589 & \textbf{0.3356} \\
\hline

\multirow{4}{*}{\makecell{Na\"{\i}ve\\Bayes}}
& AUROC (P)        & 0.9249 & 0.8655 & 0.5064 & 0.5941 & 0.8392 & 0.9475 & 0.9411 & 0.9411 & \textbf{0.9820} \\
& FRR@95\%TRR (P)  & 0.4598 & 0.7081 & 1.0000 & 1.0000 & 0.4742 & 0.4504 & 0.4681 & 0.4681 & \textbf{0.1507} \\
& AUROC (N)        & 0.7277 & 0.6332 & 0.5249 & 0.6287 & NA    & 0.6105 & 0.5531 & 0.5531 & \textbf{0.8638} \\
& FRR@95\%TRR (N)  & 0.9055 & 0.8084 & 1.0000 & 0.9271 & NA    & 0.9552 & 0.9838 & 0.9838 & \textbf{0.4842} \\
\hline
\end{tabular}

\vspace{1mm}
\raggedright\scriptsize
CFR = Classifier; UQ-D = Uncertainty Quantification Detector; Ori.=Original; Ale.=Aleatory; Vac.=Vacuity; Diss.=Dissonance; $D_\alpha$/$D_\beta$ = DOCTOR variants; P = spam; N = non-spam. NA indicates no group-level topic profile due to classifier performance.
\end{table}

\subsection{Performance Analysis of X-MAP as a Misclassification Detector} \noindent Table~\ref{tab:auroc_and_frr_topic_specific_uncertainty} summarizes the performance of X-MAP and three UQ detectors (DOCTOR, ODIN, REL-U) on the SMS dataset. X-MAP is evaluated using the original topic distribution and seven topic-level uncertainty variants. UQ detectors operate on output probabilities and, in this binary setting, DOCTOR, ODIN, and REL-U yield identical performance, so we report them collectively. Because X-MAP compares positive predictions to TP profiles and negative predictions to TN profiles, it produces two decision thresholds; correspondingly, UQ detectors are also evaluated separately for positive and negative prediction subsets.

Overall, X-MAP is particularly effective at detecting misclassified positive predictions (spam/phishing). For LogReg, the aleatoric uncertainty achieves an AUROC of 0.9763 and FRR@95\%TRR of 0.0895 on positive predictions, outperforming UQ detectors. This suggests that topic-level patterns derived from SHAP explanations are especially informative for distinguishing true from false alarms (TP vs. FP). For \emph{negative} predictions (legitimate), UQ detectors tend to perform better. This is consistent with the intuition that many false negatives do not always present strong, coherent spam-like signals.

Regarding the choice of topic representation, we observe that topic-specific aleatoric uncertainty yields the best performance on positive predictions, mainly because it captures ambiguity among semantically similar suspicious topics and is less biased by highly spammy features. For negative predictions, the original topic distribution indicates that simple deviations from the typical legitimate-topic profile (TN) are already informative; Additional uncertainty transformations sometimes obscure this signal.

Finally, even X-MAP does not surpass UQ-based detectors numerically, it provides human-interpretable, topic-level explanations for why a sample is flagged as risky, which UQ scores alone cannot offer.

\subsection{Performance  Analysis of X-MAP on Falsely Rejected Message} \noindent We next evaluate X-MAP on messages that were \emph{falsely rejected} by UQ detectors. For each UQ method in Section~\ref{sec: repair layer design}, we collect the false rejection set when base detectors get 95\% TRR. Since all UQ methods operate on the same scalar model output and the task is binary, they produce almost identical rejection sets. We therefore report results using REL-U as the base detector.

\begin{table}[t]
\centering
\caption{Performance of X-MAP on Messages Falsely Rejected by UQ-Based Detector}
\label{tab:auroc_and_frr_on_UQ_false_reject}
\scriptsize
\setlength{\tabcolsep}{2pt}
\begin{tabular}{|c|l|c|c|c|c|c|c|c|c|}
\hline
\multirow{2}{*}{\textbf{CFR}}
& \multirow{2}{*}{\textbf{Metric}}
& \multicolumn{8}{c|}{\textbf{Topic Representation}} \\
\cline{3-10}
& & Ori. & Ale. & Vac. & Diss. & REL-U & $D_\alpha$ & $D_\beta$ & ODIN \\
\hline

\multirow{4}{*}{LogReg}
& AUROC (P)        & 0.9504 & 0.9523 & 0.9470 & 0.9474 & 0.8283 & \textbf{0.9728} & 0.8545 & 0.8545 \\
& FRR@95\%TRR (P)  & 0.3066 & \textbf{0.1679} & 0.3923 & 0.2301 & 0.2321 & 0.2162 & 0.6054 & 0.6054 \\
& AUROC (N)        & 0.9471 & \textbf{0.9615} & 0.8802 & 0.8711 & 0.5252 & 0.8771 & 0.8041 & 0.8041 \\
& FRR@95\%TRR (N)  & 1.0000 & 1.0000 & \textbf{0.3574} & 0.3971 & 0.9397 & 0.3905 & 0.5515 & 0.5515 \\
\hline

\multirow{4}{*}{SVM}
& AUROC (P)        & 0.9454 & 0.9000 & 0.9143 & 0.9184 & 0.9251 & \textbf{0.9471} & 0.9049 & 0.9049 \\
& FRR@95\%TRR (P)  & \textbf{0.1856} & 0.7479 & 0.2438 & 0.2909 & \textbf{0.1856} & 0.3463 & 0.4737 & 0.3463 \\
& AUROC (N)        & 0.8194 & 0.6150 & 0.8043 & \textbf{0.8254} & 0.6134 & 0.7761 & 0.6372 & 0.6372 \\
& FRR@95\%TRR (N)  & 0.9136 & 0.8615 & 0.9024 & 0.8639 & 0.9979 & \textbf{0.5467} & 0.8327 & 0.8327 \\
\hline

\multirow{4}{*}{\makecell{Na\"{\i}ve\\Bayes}}
& AUROC (P)        & 0.9488 & 0.7810 & 0.5017 & 0.5119 & 0.9158 & \textbf{0.9544} & 0.9371 & 0.9371 \\
& FRR@95\%TRR (P)  & 0.2151 & 0.9589 & 1.0000 & 1.0000 & 0.6170 & \textbf{0.1915} & 0.2695 & 0.2695 \\
& AUROC (N)        & \textbf{0.7339} & 0.6761 & 0.5719 & 0.6939 & NA & 0.6651 & 0.6161 & 0.6161 \\
& FRR@95\%TRR (N)  & \textbf{0.8551} & 0.9194 & 1.0000 & 0.9098 & NA & 0.8957 & 0.8902 & 0.8902 \\
\hline

\end{tabular}

\vspace{1mm}
\raggedright\scriptsize
CFR = Classifier; Ori.=Original; Ale.=Aleatory; 
Vac.=Vacuity; Diss.=Dissonance; $D_\alpha$/$D_\beta$ = DOCTOR variants;  
P = spam; N = non-spam. NA indicates no group-level topic profile due to classifier performance.
\end{table}

Table~\ref{tab:auroc_and_frr_on_UQ_false_reject} shows that X-MAP maintains strong AUROC on these already ``hard’’ messages, achieving performance close to that on the full test set (Table~\ref{tab:auroc_and_frr_topic_specific_uncertainty}). This demonstrates that X-MAP captures information complementary to output-level uncertainty: topic patterns still effectively separate correctly classified samples from misclassified ones that UQ-based detectors falsely reject.

In contrast, FRR@95\%TRR is less stable on false-rejection subsets and can increase sharply. Because these subsets are smaller than the full test set and contain fewer true misclassifications, even a few re-accepted errors can largely affect FRR. We therefore treat these results mainly as evidence of complementarity rather than as reliable stand-alone operating points.

\subsection{Performance Analysis of X-MAP as a Repair Layer} \noindent Table~\ref{tab:recover_and_leak} quantifies X-MAP's impact when applied as a repair layer on top of a UQ-based detector, reporting recovery rate ($RecovR$), leakage rate ($LeakR$), and absolute counts. X-MAP restores a substantial portion of falsely rejected samples while keeping leakage moderate. For instance, with LogReg and the original topic distribution, it recovers about 94\% of false rejections and leaks only about 15\%, yielding several hundred net correct fixes. These results show that X-MAP enhances effective coverage and rejection accuracy when paired with existing UQ-based detectors.

\begin{table}[t]
\centering
\caption{Performance of X-MAP on Messages Rejected by UQ-Based Detector}
\label{tab:recover_and_leak}
\scriptsize
\setlength{\tabcolsep}{2pt}
\begin{tabular}{|c|l|c|c|c|c|c|c|c|c|}
\hline
\multirow{2}{*}{\textbf{CFR}} 
& \multirow{2}{*}{\textbf{Metric}} 
& \multicolumn{8}{c|}{\textbf{Topic Representation}} \\
\cline{3-10}
& & Ori. & Ale. & Vac. & Diss. & REL-U & $D_\alpha$ & $D_\beta$ & ODIN \\
\hline

\multirow{5}{*}{LogReg} 
& RecovR              & 0.9436 & \textbf{0.9498} & 0.9404 & 0.9279 & 0.1923 & 0.2268 & 0.8694 & 0.8276 \\
& \# Recovery         & 903    & \textbf{909}    & 900    & 888    & 184    & 217    & 832    & 792    \\
& LeakR               & \textbf{0.1542} & 0.1522 & \textbf{0.1542} & 0.1522 & 0.1502 & 0.1502 & \textbf{0.1542} & 0.1502 \\
& \# Leakage          & \textbf{78}     & 77     & \textbf{78}     & 77     & 76     & 76     & \textbf{78}     & 76     \\
& \# Correct Fix      & 825    & \textbf{832}    & 822    & 811    & 108    & 141    & 754    & 716    \\
\hline

\multirow{5}{*}{SVM} 
& RecovR              & 0.8612 & \textbf{0.9100} & 0.8903 & 0.8586 & 0.8743 & 0.8566 & 0.8527 & 0.8583 \\
& \# Recovery         & 2631   & \textbf{2780}   & 2720   & 2623   & 2671   & 2617   & 2605   & 2622   \\
& LeakR               & \textbf{0.5319} & 0.4308 & \textbf{0.5319} & 0.5160 & 0.2074 & 0.5000 & 0.4894 & 0.5213 \\
& \# Leakage          & \textbf{100}    & 81     & \textbf{100}    & 97     & 39     & 94     & 92     & 98     \\
& \# Correct Fix      & 2531   & \textbf{2699}   & 2620   & 2526   & 2632   & 2523   & 2513   & 2524   \\
\hline

\multirow{5}{*}{\makecell{Na\"{\i}ve\\Bayes}}
& RecovR              & 0.9482 & \textbf{0.9769} & 0.9607 & 0.9604 & 0.1348 & 0.9722 & 0.8552 & 0.9707 \\
& \# Recovery         & 3038   & \textbf{3130}   & 3078   & 3077   & 432    & 3115   & 2740   & 3110   \\
& LeakR               & \textbf{0.7615} & 0.6832 & 0.6868 & 0.7367 & 0.3772 & 0.5445 & 0.5907 & 0.5125 \\
& \# Leakage          & \textbf{214}    & 192    & 193    & 207    & 106    & 153    & 166    & 144    \\
& \# Correct Fix      & 2824   & 2938   & 2885   & 2870   & 326    & \textbf{2962}   & 2574   & 2866   \\
\hline
\end{tabular}

\vspace{2mm}
\raggedright\scriptsize
CFR = Classifier; Ori.=Original; Ale.=Aleatory; 
Vac.=Vacuity; Diss.=Dissonance; $D_\alpha$/$D_\beta$ = DOCTOR variants.  
\# Recovery = number of false rejections recovered by X-MAP;  
\# Leakage = misclassified samples mistakenly re-accepted;  
\# Correct Fix = \# Recovery $-$ \# Leakage.  
NA indicates no group-level topic profile due to classifier performance.
\end{table}

\section{Conclusions \& Future Work}

This paper introduced X-MAP, an explainable Misclassification Analysis and Profiling framework for spam and phishing detection. X-MAP moves beyond scalar uncertainty scores by decomposing model behavior into topic-level patterns derived from SHAP explanations. By constructing group-level topic profiles for reliably classified TP and TN messages and measuring a message's deviations from these profiles, X-MAP provides both a quantitative indicator of misclassification and an interpretable view of underlying feature behavior. The resulting topic-level explanations reveal semantically coherent patterns that can support downstream tasks such as feature engineering, data curation, and human-centered alert design.

\noindent \textbf{Key Findings.}  
Empirical results validate X-MAP’s distance-based assumption: misclassified messages exhibit substantially larger JS divergence, often 2 times, even 10 times higher, than correctly classified ones. As a stand-alone detector, X-MAP achieves competitive or superior performance to UQ-based methods, reaching up to 0.98 AUROC and reducing FRR at 95\% TRR to around 9\% on positive predictions. When used as a repair layer on top of UQ-based detectors such as DOCTOR, ODIN, and REL-U, X-MAP recovers a large fraction of falsely rejected but correct predictions while maintaining low leakage, thereby improving effective coverage without compromising reliability.

\noindent \textbf{Limitations and Future Work.}  
While X-MAP provides interpretable topic-level insights, it currently depends on SHAP explanations and NMF-based topic modeling, which may introduce computational overhead or sensitivity to feature sparsity. Future work will extend X-MAP to more complex or multimodal models to improve scalability and robustness; Incorporate various forms of uncertainty and belief modeling at topic level to enhance both interpretability and detection power; Integrate X-MAP into interactive interfaces and conducting user studies with security analysts would assess the practical value of topic-level explanations; Finally, apply X-MAP to other high-stakes domains such as intrusion detection, fraud detection, and medical decision support would help demonstrate its generality as an explainable framework for misclassification analysis.

%
%
%
%
\printbibliography

@inproceedings{liang2017odin,
  title={Enhancing the Reliability of Out-of-distribution Image Detection in Neural Networks},
  author={Liang, Shiyu and Li, Yixuan and Srikant, R},
  booktitle={Proceedings of the International Conference on Learning Representations (ICLR)},
  year={2018}
}

@book{murphy2012mlpp,
  title={Machine Learning: A Probabilistic Perspective},
  author={Murphy, Kevin P.},
  year={2012},
  publisher={MIT Press}
}

@misc{sms_spam_collection_228,
  author = {Almeida, Tiago and Hidalgo, Jos},
  title = {{SMS Spam Collection}},
  year = {2011},
  howpublished = {UCI Machine Learning Repository},
  note = {{DOI}: https://doi.org/10.24432/C5CC84}
}

@inproceedings{SHAP,
 author = {Lundberg, Scott M and Lee, Su-In},
 booktitle = {Advances in Neural Information Processing Systems},
 
 pages = {4765-4774},
 publisher = {Curran Associates, Inc.},
 title = {A Unified Approach to Interpreting Model Predictions},
 volume = {30},
 year = {2017}
}

@article{LingEmail,
  title={A memory-based approach to anti-spam filtering for mailing lists},
  author={Sakkis, Georgios and Androutsopoulos, Ion and Paliouras, Georgios and Karkaletsis, Vangelis and Spyropoulos, Constantine D and Stamatopoulos, Panagiotis},
  journal={Information retrieval},
  volume={6},
  number={1},
  pages={49--73},
  year={2003},
  publisher={Springer}
}

@article{NigerianFraud,
  title={Clair collection of fraud email, acl data and code repository},
  author={Radev, Dragomir},
  journal={ADCR2008T001},
  volume={5},
  number={5.5},
  pages={1},
  year={2008}
}

@inproceedings{hendrycks2017baseline,
  title={A Baseline for Detecting Misclassified and Out-of-Distribution Examples in Neural Networks},
  author={Hendrycks, Dan and Gimpel, Kevin},
  booktitle={International Conference on Learning Representations},
  year={2017}
}

@inproceedings{qiu2022detecting,
  title={Detecting misclassification errors in neural networks with a gaussian process model},
  author={Qiu, Xin and Miikkulainen, Risto},
  booktitle={Proceedings of the AAAI Conference on Artificial Intelligence},
  volume={36},
  number={7},
  pages={8017--8027},
  year={2022}
}

@article{granese2021doctor,
  title={Doctor: A simple method for detecting misclassification errors},
  author={Granese, Federica and Romanelli, Marco and Gorla, Daniele and Palamidessi, Catuscia and Piantanida, Pablo},
  journal={Advances in Neural Information Processing Systems},
  volume={34},
  pages={5669--5681},
  year={2021}
}

@article{klie2023annotation,
  title={Annotation error detection: Analyzing the past and present for a more coherent future},
  author={Klie, Jan-Christoph and Webber, Bonnie and Gurevych, Iryna},
  journal={Computational Linguistics},
  volume={49},
  number={1},
  pages={157--198},
  year={2023},
  publisher={MIT Press One Broadway, 12th Floor, Cambridge, Massachusetts 02142, USA~…}
}

@article{du2025balancing,
  title={Balancing misclassification errors in image-based inference using problem domain semantics and a nested cascade architecture},
  author={Du, Xin and Jena, Rajesh and Farrahi, Katayoun and Niranjan, Mahesan},
  journal={Neural Computing and Applications},
  pages={1--16},
  year={2025},
  publisher={Springer}
}

@inproceedings{spackman2023identifying,
  title={Identifying and mitigating misclassification: A case study of the Machine Learning lifecycle in price indices with web-scraped clothing data},
  author={Spackman, William and DeVilliers, Greg and Ritter, Christian and Goussev, Serge},
  booktitle={UNECE Meeting of the Group of Experts on Consumer Price Indices},
  pages={7--9},
  year={2023}
}

@inproceedings{zhu2023openmix,
  title={Openmix: Exploring outlier samples for misclassification detection},
  author={Zhu, Fei and Cheng, Zhen and Zhang, Xu-Yao and Liu, Cheng-Lin},
  booktitle={Proceedings of the IEEE/CVF conference on computer vision and pattern recognition},
  pages={12074--12083},
  year={2023}
}

@inproceedings{vazhentsev2022uncertainty,
  title={Uncertainty estimation of transformer predictions for misclassification detection},
  author={Vazhentsev, Artem and Kuzmin, Gleb and Shelmanov, Artem and Tsvigun, Akim and Tsymbalov, Evgenii and Fedyanin, Kirill and Panov, Maxim and Panchenko, Alexander and Gusev, Gleb and Burtsev, Mikhail and others},
  booktitle={Proceedings of the 60th Annual Meeting of the Association for Computational Linguistics (Volume 1: Long Papers)},
  pages={8237--8252},
  year={2022}
}

@misc{alyahya2024toward,
  title={Toward Reducing IDS Misclassification Using Hybrid DL and ML Approach. Advances in Artificial Intelligence and Machine Learning. 2024; 4 (3): 161},
  author={Alyahya, Mohammed and others},
  journal={Reliance},
  volume={9},
  pages={11},
  year={2024}
}

@article{naser2025failure,
  title={From failure to fusion: A survey on learning from bad machine learning models},
  author={Naser, MZ},
  journal={Information Fusion},
  volume={120},
  pages={103122},
  year={2025},
  publisher={Elsevier}
}

@techreport{FBI2024IC3Report,
  author       = {{FBI}},
  title        = {2023 Internet Crime Report},
  institution  = {Internet Crime Complaint Center (IC3)},
  year         = {2024},
  url          = {https://www.ic3.gov/Media/PDF/AnnualReport/2023_IC3Report.pdf},
}

@misc{FTC2025TextScams2024,
  author       = {{Federal Trade Commission}},
  title        = {New FTC Data Show Top Text Message Scams of 2024; Overall Losses to Text Scams Hit \$470 Million},
  year         = {2025},
  note         = {\href{https://www.ftc.gov/news-events/news/press-releases/2025/04/new-ftc-data-show-top-text-message-scams-2024-overall-losses-text-scams-hit-470-million}{FTC press release}},
}

@inproceedings{gutierrez2020email,
  title={Email embeddings for phishing detection},
  author={Guti{\'e}rrez, Luis Felipe and Abri, Faranak and Armstrong, Miriam and Namin, Akbar Siami and Jones, Keith S},
  booktitle={2020 ieee international conference on big data (big data)},
  pages={2087--2092},
  year={2020},
  organization={IEEE}
}

@incollection{nair2025phishemailllm,
  title={PhishEmailLLM: A Meta Model Approach to Detect Phishing emails by leveraging LLMs and Machine Learning models},
  author={Nair, Ronish and Abbasi, Fahim and Pervez, Shahbaz},
  booktitle={Proceedings of the 2025 Australasian Computer Science Week},
  pages={19--29},
  year={2025}
}

@article{lin2002divergence,
  title={Divergence measures based on the Shannon entropy},
  author={Lin, Jianhua},
  journal={IEEE Transactions on Information theory},
  volume={37},
  number={1},
  pages={145--151},
  year={2002},
  publisher={IEEE}
}

@INPROCEEDINGS{Josang18,
  author={Josang, Audun and Cho, Jin-Hee and Chen, Feng},
  booktitle={2018 21st International Conference on Information Fusion (FUSION)}, 
  title={Uncertainty Characteristics of Subjective Opinions}, 
  year={2018},
  volume={},
  number={},
  pages={1998-2005}}

@book{jsang2018subjective,
  title={Subjective Logic: A formalism for reasoning under uncertainty},
  author={Jsang, Audun},
  year={2018},
  publisher={Springer Publishing Company, Incorporated}
}

@inproceedings{gomes2023rel-u,
  title={A data-driven measure of relative uncertainty for misclassification detection},
  author={Dadalto, Eduardo and Romanelli, Marco and Pichler, Georg and Piantanida, Pablo},
  booktitle={The Twelfth International Conference on Learning Representations},
  year={2023}
}

@inproceedings{liang2018enhancing,
  title={Enhancing The Reliability of Out-of-distribution Image Detection in Neural Networks},
  author={Liang, Shiyu and Li, Yixuan and Srikant, R},
  booktitle={International Conference on Learning Representations},
  year={2018}
}

@inproceedings{renyi1961entropy,
  title={On measures of entropy and information},
  author={R{\'e}nyi, Alfr{\'e}d},
  booktitle={Proceedings of the fourth Berkeley symposium on mathematical statistics and probability, volume 1: contributions to the theory of statistics},
  volume={4},
  pages={547--562},
  year={1961},
  organization={University of California Press}
}

@article{lee2000nmf,
  title={Algorithms for non-negative matrix factorization},
  author={Lee, Daniel and Seung, H Sebastian},
  journal={Advances in neural information processing systems},
  volume={13},
  year={2000}
}

\end{document}